\documentclass[runningheads]{comsis2}

\def\journalissue{}
\def\paperidnum{}
\usepackage[all]{xy,xypic}
\usepackage{eufrak,amscd,bezier,latexsym,mathrsfs,eurosym,enumerate}
\usepackage{amsfonts}
\usepackage{amssymb}
\usepackage{amsmath}
\usepackage{amsgen}
\usepackage{amsopn}
\usepackage{amsbsy}
\usepackage{theorem}
\usepackage{graphicx}
\usepackage{soul}
\usepackage{epsfig}
\usepackage[utf8]{inputenc}
\usepackage[english]{babel}
\usepackage[dvipsnames]{xcolor}
\usepackage[pagewise]{lineno}
\usepackage{tipa}
\usepackage{accents}
\usepackage{adjustbox}
\usepackage{array}
\usepackage[numbers]{natbib}





\newcolumntype{C}[1]{>{\centering\let\newline\\\arraybackslash\hspace{0pt}}m{#1}}

\title{Evaluation of Phonetic Encoding Algorithms on Transcription Datasets}

\titlerunning{Evaluation of Phonetic Encoding Algorithms}

\author{Can Özbey\inst{1} \and Emre Kaplan\inst{1} \and Berkin Deniz Kahya\inst{1,2}}

\institute{Huawei Turkey R\&D Center\\
	Istanbul, Turkey\\
	\and
	Dept. of Computer Engineering, Istanbul Technical University\\
	Istanbul, Turkey\\
	\email{kahya25@itu.edu.tr}}

\begin{document}

\maketitle

\begin{abstract}

In this work, a novel evaluation scheme built on a generalized variant of the Rand Index measure, namely, the Hüllermeier-Rifqi Index, is proposed in order to assess how well phonetic encoding algorithms conform to word-based transcriptions in IPA (International Phonetic Alphabet) notation. For this objective, the discordance score is obtained by calculating the absolute difference between the pairwise similarity values of ground-truth transcriptions and those of corresponding phonetic encodings, which are computed using normalized edit distance as a permutation dependent string metric. The resulting score is subsequently adjusted with respect to that of a random string generator incorporating the same alphabet as the encoder under consideration. A wide range of phonetic encoders were evaluated as such on multi-lingual transcription datasets along with their recall capabilities based on the collision rate. The validity of the proposed scheme is further supported by its applicability in measuring the orthographic transparency of a language when the writing system is viewed as an inherent phonetic representation.

\vspace{6pt}\textbf{Keywords:} evaluation, phonetic encoding, phonetic transcription, hüllermeier-rifqi index, orthographic transparency.
\end{abstract}

\section{Introduction}

Phonetic encoding techniques provide a systematic way to translate written words into codes based on their pronunciation. The generated phonetic codes facilitate the comparison of words that are phonetically similar yet orthographically distinct. Their applications extend to diverse subdomains of computer science and information systems, including record linkage, data deduplication and normalization, phonetic search, spelling correction, transliteration, string similarity measurement, and other related areas \cite{overview}. Training neural networks on phonetic rather than orthographic representations, e.g., using pinyin for Chinese \cite{nn2} or phonetic encoders for English \cite{nn1}, has been shown to empirically improve performance in various downstream natural language processing tasks such as language modeling, machine translation and part-of-speech tagging. The application of edit distance on phonetic encodings is also a common practice to measure similarity between speech transcriptions with the aim of compensating for errors generated by speech recognition systems \cite{asr0, asr1, asr2, asr3}.

While the first formal account of a phonetic encoding procedure for name matching appears in a 1918 patent by Robert C. Russell \cite{russell}, a conceptual antecedent can be traced back at least to the mid-19th century, notably in the work of Beniowski (1845) \cite{phrenotype}, who had proposed a mnemonic technique for effective number memorization by transforming numbers into words. Even though these two methods serve different purposes, Beniowski’s method of decoding words to recall numbers employs a procedure notably similar to that of Russell's phonetic encoding, as both omit vowels and rely on mapping individual consonants or consonant groups to digits. Their main distinction lies in the direction in which encoding and decoding are carried out, that is, Russell's method is based on many-to-one mapping of letters to digits to obtain phonetic representation of words, whereas Beniowski's method is based on one-to-many mapping of digits to letters. Thus, the decoding phase of the latter is in fact structurally equivalent to the encoding phase of the former.

One of the best-known phonetic encoding algorithms, American Soundex \cite{soundex}, was later developed as a variant of Russell's encoding with several important modifications, including the retention of the first letter and the prefix truncation of the resulting code to a fixed length of characters. Six letter clusters were defined as groups of consonants organized by their broad phonetic similarity to construct the mapping rules, namely, \{\textit{b, f, p, v}\}, \{\textit{c, g, j, k, q, s, x, z}\}, \{\textit{d, t}\}, \{\textit{l}\}, \{\textit{m, n}\}, and \{\textit{r}\}. As such, each consonant is converted into a corresponding digit assigned to its cluster, with the exception of the first letter, which is preserved regardless of whether it is a vowel or a consonant. Several extensions of this algorithm, such as Refined Soundex \cite{refinedsoundex} and Fuzzy Soundex \cite{fuzzysoundex}, were later developed to provide finer granularity by means of extended consonant groups, while also allowing flexibility in code length. Metaphone \cite{metaphone}, in particular, introduced a significantly more fine-grained encoding methodology through a comprehensive list of transformation rules that also involve multiple letter sequences denoting digraphs, being designed to better account for the highly irregular nature of English pronunciation. Its successor, Double Metaphone \cite{dmetaphone}, extended the scope even further by incorporating additional rules for handling foreign words borrowed from other languages into English such as surnames derived from Slavic, Germanic, Romance, and several other language groups.

Evaluating phonetic encoders typically involves measuring how well they identify spelling variants or similar sounding words, particularly of proper nouns \cite{phonex, name}, and as well as their performance in detecting duplicate entries in data storage systems \cite{duplicate}. Several studies have examined the impact of phonetic encoding algorithms on the performance of automatic spelling correction systems \cite{sc1, sc2}, since employing phonetic information has a potential to improve the effectiveness of correcting orthographic errors \cite{sc3}. An explicit example of such an evaluation was provided in \cite{soundexgr}, where a Soundex-like phonetic encoding technique designed specifically for Greek was evaluated with a focus on its performance in correcting spelling errors generated via edit operations. Notably, the study further demonstrated that even a full phonetic transcription may be outperformed by a well-designed phonetic code in retrieving similar sounding words resulting from the latter’s higher recall rate. Thus, recall capability constitutes another key factor in the evaluation of phonetic encoders as highlighted by several other previous studies \cite{recall1, recall2}, which suggest that phonetic codes with low granularity may be preferable depending on the task. The degree of granularity can be effectively measured by simply computing the collision rate \cite{r}, i.e., the average number of shared codes per input, reflecting an encoder’s potential recall capability. However, to the best of our knowledge, no systematic evaluation methodology focusing on precision has been presented in the literature, aside from the extrinsic approaches mentioned earlier. Our objective in this work is to provide an evaluation scheme where we formulate a measure of the consistency between phonetic codes and ground-truth transcriptions, allowing for a direct assessment of how accurately the codes align with the actual sequence of sounds. In combination with the collision rate analysis, such an evaluation is fundamental to establishing a robust framework for thoroughly understanding the general capability of phonetic encoders. The proposed approach will make it feasible to conduct a comparative analysis based on the consistency measure between codes and transcriptions for those having a similar collision rate, and vice versa, in order to identify the better-performing one without depending on their specific performance metrics in downstream tasks.

To establish the mathematical basis of the evaluation scheme, we adapt the Hüllermeier-Rifqi Index, i.e., a generalized variant of the Rand Index, itself originally formulated to evaluate predictions of clustering algorithms. Given that both orthographic forms and transcriptions are permutation-dependent, we incorporate normalized edit distance into the computation of pairwise similarities for deriving the concordance score. Furthermore, sample pairs are partitioned into positive and negative sets of equal cardinality, thereby ensuring reliable evaluation and enabling the derivation of fuzzy counterparts of conventional measures defined in a discrete setting, namely, positive, negative, and overall concordance, corresponding to sensitivity, specificity, and accuracy, respectively.  In the subsequent section, we first elaborate on how the concordance measure of Hüllermeier is formally modified to align with our objective. Then, in section 3, we apply the evaluation scheme to transcription datasets in four different languages (English, German, French and Swedish) using a diverse set of language-specific phonetic encoders, the performance of which is investigated based on their adjusted overall concordance scores and collision rates.  At the end of the section, we show that the proposed scheme can be applied to measure the orthographic transparency of a language by regarding the writing system as its inherent phonetic representation, thus providing additional validation of the approach’s effectiveness. Lastly, in section 4, we provide concluding remarks together with several suggestions for further improvement.

\section{Methodology}

\subsection{Rand Index}

The Rand Index (RI) is a measure to evaluate clustering algorithms by calculating the number of pairwise agreements between the predicted labels and the ground-truth labels \cite{rand}. Let $\mathbb{S}=\{X_1, X_2,...,X_n\}$ be a set consisting of $n$ samples. We consider $\mathbb{P}=\{P_1,...,P_r\}$ to be a partition of $\mathbb{S}$, where $P_i$ denotes a subset of $\mathbb{S}$, any of whose elements may be interpreted as being assigned to label $i$ by a clustering procedure. Similarly, let $\mathbb{Q}=\{Q_1,,...,Q_k\}$ be defined according to the partitioning induced by the given ground-truth labels. Then, a pair of two samples $\left (x, y\right )$ is said to be concordant if and only if their grouping (either together or apart) is the same in both $\mathbb{P}$ and $\mathbb{Q}$. More specifically, the pair $\left (x, y\right )$ is positively concordant when \eqref{cond1} holds, and negatively concordant when \eqref{cond2} holds. Hence, it is concordant if either of these conditions holds as expressed in \eqref{conc}.

\begin{equation}
	\left (x \in P_i\ \land\ y \in P_i\right )\ \land\ \left (x \in Q_j\ \land\ y \in Q_j\right )
	\label{cond1}
\end{equation}

\begin{equation}
	\left (x \in P_i\ \land\ y \notin P_i\right )\ \land\ \left (x \in Q_j\ \land\ y \notin Q_j\right )
	\label{cond2}
\end{equation}

\begin{align}
	C(x,y) = \begin{cases}
		1 & \text{if \eqref{cond1} $\lor$ \eqref{cond2}} \\
		0 & \text{otherwise}
	\end{cases}\label{conc}
\end{align}

The RI is computed by dividing the number of all concordant pairs by the total number of pairwise combinations. Let $\mathbb{C}=\left\{\left (X_i, X_j\right ) \in \mathbb{S}\times\mathbb{S}\mid 1\leq i<j\leq n\ \land\ C\left (X_i,X_j\right )\right\}$ denote the set of all concordant pairs. Thus, it can be expressed as the ratio given in \eqref{rand}.

\begin{equation}
	RI=\frac{\left |\mathbb{C}\right |}{{n\choose 2}}
	\label{rand}
\end{equation}

\subsection{Hüllermeier-Rifqi Index}

One of the earliest adoptions of the RI to fuzzy partitions was provided by Campello (2007) within an analytical framework \cite{campello}, where a fuzzy partition $\mathbb{P}$ is evaluated against a non-fuzzy ground-truth partition $\mathbb{Q}$. Later, Hüllermeier et al. (2012) developed a more generalized variant \cite{rifqi}, establishing its validity as a proper metric, and also demonstrating that it can be used to compare two fuzzy partitions. To compute the Hüllermeier-Rifqi Index (HRI), an equivalence relation is first defined over a given pair of samples $\left (x, y\right )$ as expressed in \eqref{eq}, where $P(X)$ represents a function that yields a fuzzy membership vector of a sample across all partitions. Thus, $E\left (x, y\right )$ can be interpreted as a function for measuring similarity between two samples based on the distance between their membership vectors in a partition space. 

\begin{equation}
	E\left (x, y\right )=1 - \left\Vert P(x)-P(y) \right\Vert
	\label{eq}
\end{equation}

Analogous to the concordance function in \eqref{conc}, a discordance function $D(x,y)$ is defined as given in \eqref{disc}, where $C(x,y)=1-D(x,y)$ also holds provided that $0 \leq D(x,y) \leq 1$. 

\begin{equation}
	D\left (x, y\right )=\left\vert E_P\left (x, y\right )-E_Q\left (x, y\right ) \right\vert
	\label{disc}
\end{equation}

Here, the discordance score between two samples is the absolute difference of their pairwise similarity values in the partition spaces $\mathbb{P}$ and $\mathbb{Q}$, obtained respectively by \eqref{eq}. The discordance reaches its minimum at 0, and concordance its maximum at 1, when $E_P\left (x, y\right )=1$ and $E_Q\left (x, y\right )=1$, or when $E_P\left (x, y\right )=0$ and $E_Q\left (x, y\right )=0$, which in fact correspond to the logical conditions previously stated in \eqref{cond1} and \eqref{cond2} for the computation of the RI.

To calculate the HRI score, the discordance values are summed over the set of all pairwise combinations, which is denoted by $\mathbb{U}=\left\{\left (X_i, X_j\right ) \in \mathbb{S}\times\mathbb{S}\mid 1\leq i<j\leq n\right\}$, and the resulting ratio is subtracted from 1, as shown in \eqref{hri}.

\begin{equation}
	HRI=1 - \frac{\sum_{(x,y)\in\mathbb{U}} D\left (x, y\right )}{{n\choose 2}}
	\label{hri}
\end{equation}

\subsection{Selection of the Distance Metric}

The HRI was originally formulated to compare membership vectors without sensitivity to order, that is, it yields the same score regardless of how the labels are arranged. However, string comparisons, such as in our case, are order-dependent. The HRI can be modified to become order-dependent by selecting a suitable distance metric $\left ( \left\Vert\cdot\right\Vert\right )$ for the similarity measurement in \eqref{eq}. As shown in \eqref{dist}, we incorporate normalized edit distance, which supports insertion, deletion and substitution to measure similarity between a given pair of strings, $x$ and $y$.

\begin{equation}
	E\left (x, y\right )=1 - \frac{d_{edit}\left (x, y\right )}{\max\left (|x|, |y|\right )}
	\label{dist}
\end{equation}

For instance, let us consider two English words ``\textit{bracing}" and ``\textit{breaking}" with their phonetic transcriptions \textipa{/}\textit{\textipa{b}\textturnr\textipa{e}\textsci\textipa{s}\textsci\textipa{\ng}}\textipa{/} and \textipa{/}\textit{\textipa{b}\textturnr\textipa{e}\textsci\textipa{k}\textsci\textipa{\ng}}\textipa{/}, respectively. The similarity scores are calculated as $E_P\left (x, y\right )=\frac{3}{4}$ and $E_Q\left (x, y\right )=\frac{6}{7}$, the latter referring to the transcription pair. Thus, the computed discordance score according to \eqref{disc} equals $\frac{3}{28}\approx 0.107$. As another example, let us also consider ``\textit{oceans}" and ``\textit{motions}" with their phonetic transcriptions \textipa{/}\textit{\textipa{o}\textupsilon\textesh\textschwa\textipa{n}\textipa{z}}\textipa{/} and \textipa{/}\textit{\textipa{m}\textipa{o}\textupsilon\textesh\textschwa\textipa{n}\textipa{z}}\textipa{/}, respectively. Despite $E_Q\left (x, y\right )=\frac{6}{7}$, as in the former case, $E_P\left (x, y\right )=\frac{3}{7}$ resulting in four times higher discordance score of $\frac{3}{7}\approx 0.429$. This suggests that, relative to the former example, the orthographic representations of the latter deviate from each other considerably more than that implied by their phonetic counterparts.

\subsection{Reformulation of the Index}

The HRI computes the total discordance score by enumerating across all pairwise combinations denoted by $\mathbb{U}$ previously. In our case, however, this would lead to an overestimation of the index due to the predominance of low-similarity pairs. Therefore, we divide the complete set $\mathbb{U}$ into positive and negative subsets, aiming for comparable cardinalities in order to conduct a more reliable evaluation. Pairs whose similarity values in the ground-truth partition space are greater than a predetermined cutoff threshold $\theta^{+}\in [0, 1)$, are included in the positive set as expressed in \eqref{pos}. The pairs given as examples in the previous subsection may qualify as positive samples since their similarity scores between the phonetic transcriptions are quite high: $E_Q\left (x, y\right )=\frac{6}{7}$.

\begin{equation}
	\mathbb{U}^{+}=\left\{\left (X_i, X_j\right ) \in \mathbb{U}\mid E_Q\left (X_i,X_j\right )>\theta^{+}\right\}
	\label{pos}
\end{equation}

In a similar manner, we also collect pairs whose similarity values are less than or equal to a cutoff threshold $\theta^{-}$ provided that $0\leq\theta^{-}\leq\theta^{+}$, and include them in the negative set as expressed in \eqref{neg}. Maintaining similar cardinalities for $\mathbb{U}^{+}$ and $\mathbb{U}^{-}$ is preferred. Yet, under typical conditions, $\left\vert\mathbb{U}^{-}\right\vert$ is expected to exceed $\left\vert\mathbb{U}^{+}\right\vert$ due to the prevalence of low-similarity pairs unless $\theta^{+}$ is assigned an unreasonably low value\footnote{In our setting, these parameters were set as $\theta^{+}=0.5$ and $\theta^{-}=0$.}. Thus, a further random sampling from $\mathbb{U}^{-}$ is likely to be necessary to equalize its cardinality with that of $\theta^{+}$.

\begin{equation}
	\mathbb{U}^{-}=\left\{\left (X_i, X_j\right ) \in \mathbb{U}\mid E_Q\left (X_i,X_j\right )\leq\theta^{-}\right\}
	\label{neg}
\end{equation}

Additionally, the discordance function in \eqref{disc} was also modified and applied to the samples in $\mathbb{U}^{+}$ and $\mathbb{U}^{-}$ differently. As shown in \eqref{posdisc}, the discordance score is computed as nonzero only if $E_Q\left (x, y\right )>E_P\left (x, y\right )$. for the samples in $\mathbb{U}^{+}$. The earlier examples comply with this condition in that $E_P\left (x, y\right )<\frac{6}{7}$ yielding nonzero scores of $0.107$ and $0.429$.

\begin{equation}
	D^{+}\left (x, y\right )=\max\bigl\{ 0,\ E_Q\left (x, y\right )-E_P\left (x, y\right )\bigl\}
	\label{posdisc}
\end{equation}

Discordance in the case of positive samples may be interpreted as a false negative in a conventional classification scenario. Thus, we propose positive concordance (PC) as a fuzzy extension of the \textit{sensitivity} (recall) measure as expressed in \eqref{posconcord}, where the number of true positives is $\left\vert\mathbb{U}^{+}\right\vert$.

\begin{equation}
	PC = \frac{\left\vert\mathbb{U}^{+}\right\vert}{\left\vert\mathbb{U}^{+}\right\vert + \sum_{(x,y)\in\mathbb{U^{+}}} D^{+}\left (x, y\right )}
	\label{posconcord}
\end{equation}

The discordance function was adapted also for the negative samples in a similar manner as shown in \eqref{negdisc}. In this case, the score is computed as nonzero if $E_P\left (x, y\right )>E_Q\left (x, y\right )$. As an example, consider two English words ``\textit{leary}" and ``\textit{hearn}" with their phonetic transcriptions \textipa{/}\textit{\textipa{l}\textbabygamma\textsci\textturnr\textipa{i}}\textipa{/} and \textipa{/}\textit{\textipa{h}\textrhookrevepsilon\textipa{n}}\textipa{/}, respectively. The similarity scores are calculated as $E_P\left (x, y\right )=\frac{3}{5}$ and $E_Q\left (x, y\right )=0$. Since the lowest possible value of $\theta^{-}$ is 0, this sample pair would be guaranteed to be included in $\mathbb{U}^{-}$, and $D^{-}\left (x, y\right )$ is calculated as $0.6$ according to \eqref{negdisc}.

\begin{equation}
	D^{-}\left (x, y\right )=\max\bigl\{ 0,\ E_P\left (x, y\right )-E_Q\left (x, y\right )\bigl\}
	\label{negdisc}
\end{equation}

In the case of negative samples, discordance may be interpreted as a false positive. Thus, we propose negative concordance (NC) as a fuzzy extension of the \textit{specificity} measure as expressed in \eqref{negconcord}, where the number of true negatives is $\left\vert\mathbb{U}^{-}\right\vert$. 

\begin{equation}
	NC = \frac{\left\vert\mathbb{U}^{-}\right\vert}{\left\vert\mathbb{U}^{-}\right\vert + \sum_{(x,y)\in\mathbb{U^{-}}} D^{-}\left (x, y\right )}
	\label{negconcord}
\end{equation}

Combining $PC$ and $NC$, we also compute overall concordance (OC) as in \eqref{overconcord}, which corresponds to the \textit{accuracy} measure, serving as a general indicator of the performance.

\begin{equation}
	OC = \frac{\left\vert\mathbb{U}^{+}\right\vert + \left\vert\mathbb{U}^{-}\right\vert}{\frac{\left\vert\mathbb{U}^{+}\right\vert}{PC} + \frac{\left\vert\mathbb{U}^{-}\right\vert}{NC}}
	\label{overconcord}
\end{equation}

Finally, the $OC$ score is adjusted with respect to that obtained from a random model, denoted as $ROC$. As shown in \eqref{aoc}, the adjustment formulation is in fact the same as for the standard RI. To compute $ROC$, random codes of the same length as the phonetic code in each sample were generated using the same alphabet as the code alphabet, and the evaluation scheme was similarly applied to the random codes. For example, the $ROC$ score corresponding to the English writing system would involve generating random codes by sampling from the standard English alphabet, with code lengths equal to those of the orthographic representations. Equation \eqref{overconcord} yields $ROC$ after obtaining the $PC$ and $NC$ scores of random codes.

\begin{equation}
	AOC = \frac{OC - ROC}{1 - ROC}
	\label{aoc}
\end{equation}

\section{Experimental Results}

\subsection{Dataset}

We employed a multilingual phonetic transcription dataset\footnote{github.com/open-dict-data/ipa-dict} compiled from various resources, commonly used to train grapheme-to-phoneme models \cite{dict}. For the application of phonetic algorithms, only English (US), German, French, and Swedish were selected as target languages since such algorithms are designed for specific languages and are not as readily available for others. On the other hand, Dutch, Esperanto, Finnish, Icelandic, Portuguese, and Romanian were additionally evaluated solely on the basis of their writing system, and compared with the other four in that respect. The number of word-transcription pairs in the dataset vary depending on the language, ranging from a minimum of 21,108 (Swedish) to a maximum of 787,106 (German).

\subsection{Preprocessing}

The data underwent several preprocessing operations before computing the evaluation metrics. The whitespace, apostrophe and hyphen were removed from the lowercased orthographic representations, and they are excluded from the dataset if any of the remaining characters fall outside of the standard alphabet of the language. As for IPA representations, only symbols denoting vowels and consonant are retained by stripping away those indicating stress, length, breaks, prosodic features, etc. along with whitespace and hyphens. For instance, the German word-transcription pair ``\textit{achteten wert}" and \textipa{/}\textit{\textsecstress\textipa{a}\textipa{x}\textipa{t}\textschwa\textipa{t}\textsyllabic{n} \textprimstress\textipa{v}\textipa{e}\textlengthmark\textturna\textsubarch{t}}\textipa{/} would be transformed into ``\textit{achtetenwert}" and \textipa{/}\textit{\textipa{a}\textipa{x}\textipa{t}\textschwa\textipa{t}\textipa{n}\textipa{v}\textipa{e}\textturna\textipa{t}}\textipa{/}. Samples that have multiple phonetic transcriptions as well as those incorporating identical spellings but different pronunciations (heteronyms) were also excluded from the dataset.

Following the preprocessing steps, we randomly selected 10,000 samples from each language to mitigate any possible bias resulting from size differences. The sample size could not be significantly increased beyond 10,000 due to the fact that the time complexity of the evaluation scheme is quadratic.

\subsection{Evaluation}

An open-source library, namely, Abydos\footnote{Abydos (2018). NLP/IR library for Python. https://abydos.readthedocs.io/en/latest/abydos.phonetic.html}, was employed to produce phonetic codes. The orthographic representations, on which the phonetic algorithms are to be applied, were regarded as intrinsic phonetic codes in their own right, and likewise included in the evaluation scheme in order to assess how accurately they match transcriptions, indicating the level of orthographic transparency of the language. Although the vast majority of the codes are designed for English, there also exist several codes implemented for German, French, and Swedish as part of the library. The subsequent lists specify the phonetic codes evaluated for each language in addition to their own writing system.

\begin{itemize}
	\item \textbf{English}: The Alpha Search Inquiry System (AlphaSIS) code \cite{lein}, American Soundex \cite{soundex}, Caverphone \cite{caverphone}, Daitch-Mokotoff Soundex \cite{mokotoff}, Dolby \cite{dolby}, Double Metaphone \cite{dmetaphone}, Fuzzy Soundex \cite{fuzzysoundex}, Law Enforcement Information Network (LEIN) name coding \cite{lein}, Metaphone \cite{metaphone}, Match Rating Algorithm (MRA) \cite{lein}, Naval Research Laboratory (NRL) encoder \cite{nrl}, The New York State Identification and Intelligence System (NYSIIS) algorithm \cite{nysiis}, Oxford Name Compression Algorithm (ONCA) \cite{onca}, Parmar-Kumbharana \cite{parmarkumbharana}, Phonex \cite{phonex}, PHONIC \cite{nysiis}, Phonix \cite{phonix}, PSHP \cite{pshp}, Refined Soundex \cite{refinedsoundex}, Roger Root \cite{lein}, Russell Index \cite{russell}, SoundD \cite{soundd}, and Statistics Canada \cite{lein}. 
	\item \textbf{German}: Haase \cite{haase}, Kölner \cite{kolner}, Phonem \cite{phonem}, Phonet \cite{phonet}, and Reth-Schek \cite{rethschek}.
	\item \textbf{French}: FONEM \cite{fonem}, and Early Henry \cite{henry}.
	\item \textbf{Swedish}: SfinxBis \cite{sfinxbis}, and Wåhlin \cite{waahlin}.
\end{itemize}

Certain phonetic algorithms enable the specification of a maximum code length as a parameter. If exceeded, the code is truncated to fit the specified length. To facilitate an unbiased comparison across encoders, we did not impose a limit on the output code length in our experimental setting. Yet, the effect of length constraints on the trade-off between collision rate and accuracy was investigated in a separate experiment.

An overview of the results for the phonetic encoders relevant for English is provided in Table \ref{tab1} with the alphabet size $(|\Sigma|)$, the collision rate $(\mu)$, $PC$, $NC$, $OC$, $AOC$ scores as well as with their abbreviated forms to make later visuals easier to read. The writing system is also included in the evaluation, with orthographic representations regarded as inherent phonetic codes. As seen from the table, the Russell Index (RI) resulted in the highest $PC$ score of $93.93$ and the lowest $NC$ score of $79.72$, exhibiting the worst performance with an $AOC$ score of $41.02$. This indicates that RI shows high sensitivity but very low specificity, which led to a reduced overall accuracy. On the other side of the spectrum, MRA yields the highest $NC$ score of $95.36$ and a low $PC$ score of $86.05$, resulting in a comparatively decent $AOC$ score of $59.30$. Together with Parmar-Kumbharana (PK) and Statistics Canada (SC), these 3 encoders demonstrate high specificity, albeit with limited sensitivity. Notably, the highest $AOC$ is attained by Metaphone (MP) with a score of $63.39$, having a strong $NC$ $(93.57)$ without compromising $PC$ $(89.41)$. This result is not surprising, given that MP comprises diverse transformation rules specifically designed to account for the irregular pronunciation patterns of English. The two extensions of Soundex (SX), i.e., Refined Soundex (RSX) and Fuzzy Soundex (FSX), both showed improved $AOC$ scores with $58.09$ and $59.05$ over the original algorithm ($55.02$), a conceivable result that can be attributed to their more intricate phonetic mapping procedures.

\begin{table}[t]
	\caption{Evaluation of the phonetic encoders in English}
	\centering
	\setlength\tabcolsep{0.5em}
	\begin{tabular}{|c|c|c|c|c|c|c|c|}
		\hline
		\hline
		\textbf{Encoder} & \textbf{Abbr.} & $\boldsymbol{|\Sigma|}$ & $\boldsymbol{\mu}$ & $\boldsymbol{PC}$ & $\boldsymbol{NC}$ & $\boldsymbol{OC}$ & $\boldsymbol{AOC}$\\
		\hline
		\hline
		Writing System & WS & 26 & 1.00 & 84.89 & 92.46 & 88.52 & 51.08\\
		\hline
		Russell Index & RI & 8 & 7.48 & \textbf{93.93} & 79.72 & 86.24 & 41.02\\
		\hline
		Soundex & SX & 26 & 4.44 & 90.23 & 88.79 & 89.50 & 55.02\\
		\hline
		Refined Soundex & RSX & 26 & 3.27 & 89.20 & 91.25 & 90.21 & 58.09\\
		\hline
		Daitch-Mokotoff & DMF & 10 & 4.52 & 89.25 & 87.97 & 88.60 & 51.22\\
		\hline
		Metaphone & MP & 21 & 2.54 & 89.41 & 93.57 & \textbf{91.44} & \textbf{63.39}\\
		\hline
		SoundD & SDD & 6 & 8.03 & 91.94 & 84.71 & 88.18 & 48.95\\
		\hline
		Roger Root & RR & 14 & 4.41 & 88.72 & 92.74 & 90.69 & 60.10\\
		\hline
		AlphaSIS & ASIS & 14 & 4.73 & 88.21 & 92.85 & 90.47 & 59.07\\
		\hline
		Caverphone & CP & 12 & 3.91 & 89.29 & 89.29 & 89.29 & 54.26\\
		\hline
		NYSIIS & NYS & 25 & 2.39 & 92.18 & 82.68 & 87.17 & 45.15\\
		\hline
		Phonix & PHIX & 19 & 4.26 & 89.58 & 91.88 & 90.71 & 60.20\\
		\hline
		PHONIC & PHIC & 10 & 3.97 & 89.80 & 91.54 & 90.66 & 59.81\\
		\hline
		Phonex & PHEX & 15 & \textbf{11.08} & 88.47 & 89.41 & 88.94 & 52.66\\
		\hline
		LEIN & LEIN & 26 & 5.22 & 89.78 & 86.09 & 87.89 & 48.14\\
		\hline
		Dolby & DO & 17 & 3.26 & 93.51 & 82.49 & 87.65 & 47.31\\
		\hline
		Double Metaphone & DMP & 14 & 3.19 & 89.90 & 92.35 & 91.11 & 61.99\\
		\hline
		MRA & MRA & 26 & 1.98 & 86.05 & \textbf{95.36} & 90.47 & 59.30\\
		\hline
		Fuzzy Soundex & FSX & 26 & 3.65 & 89.69 & 91.15 & 90.41 & 59.05\\
		\hline
		Statistics Canada & SC & 26 & 2.08 & 86.65 & 95.22 & 90.73 & 60.38\\
		\hline
		Parmar-Kumbharana & PK & 26 & 2.05 & 87.09 & 95.20 & 90.97 & 61.40\\
		\hline
		PSHP & PSHP & 22 & 8.85 & 89.39 & 86.71 & 88.03 & 48.76\\
		\hline
		NRL & NRL & 35 & 1.11 & 86.45 & 90.94 & 88.64 & 51.59\\
		\hline
		ONCA & ONCA & 26 & 5.43 & 89.33 & 90.76 & 90.04 & 57.32\\
		\hline
	\end{tabular}
	\label{tab1}
\end{table}

\begin{figure}[!t]
	\centering
	
	\includegraphics[width=1.0\textwidth]{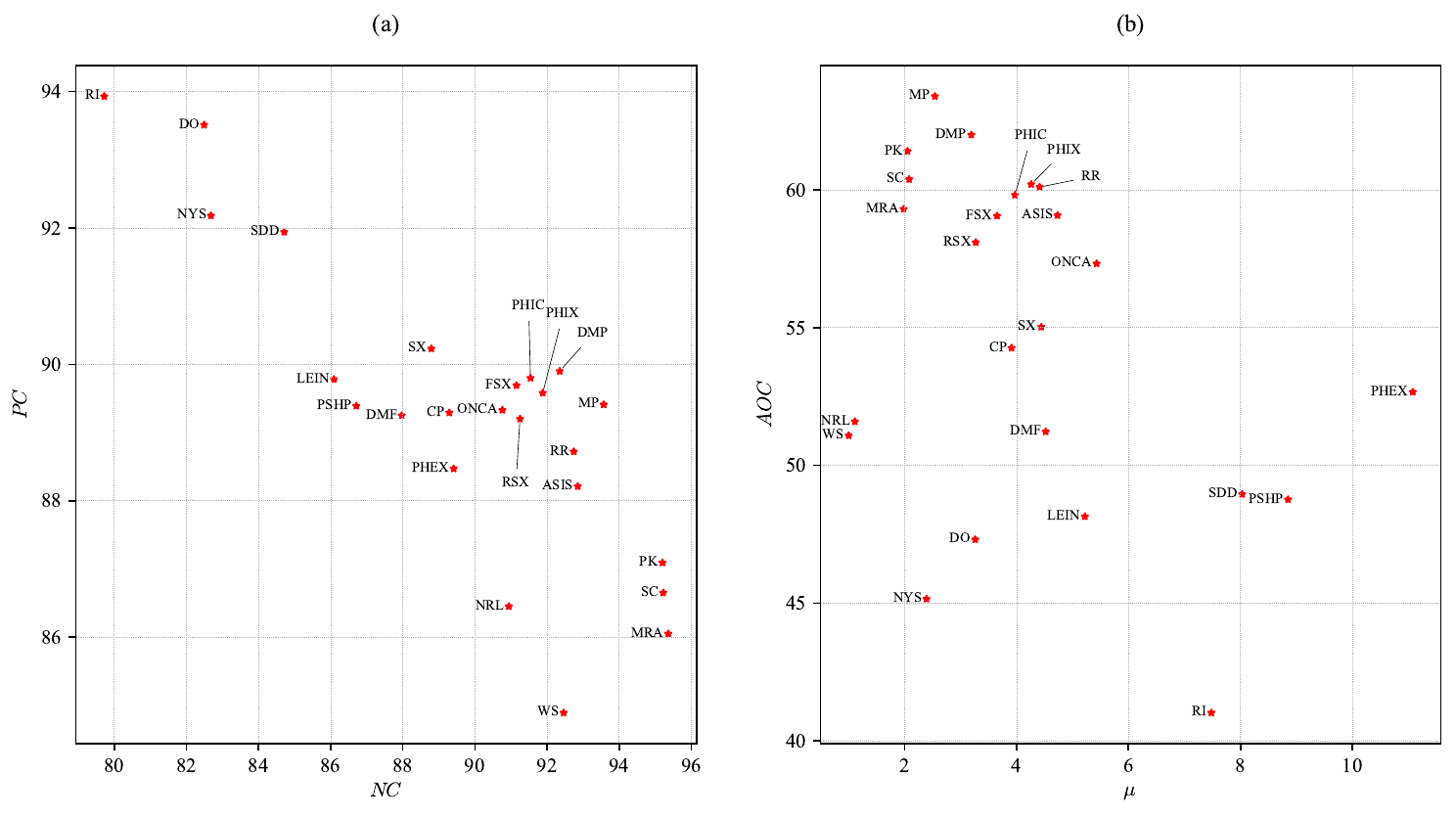}
	
	\caption{Performance of the phonetic encoders in English showing the $PC$ and $NC$ scores (left) and the $AOC$ scores with collision rates (right)}
	\label{fig1}
\end{figure}

In Figure \ref{fig1}, the results given in the table are visually depicted to better illustrate the performance of the encoders from multiple perspectives. In the left subplot (a), the $PC$ (y-axis) and the $NC$ (x-axis) scores are provided to demonstrate the trade-off between the two measures. On this spectrum, a comparison of the three methods, namely, RI, SX and MRA, is particularly informative since all of them remove vowels and differ mainly in how they handle consonant transformations. In broad terms, RI transforms all consonants to their group identifiers, SX preserves the initial consonant and applies the same procedure to the remaining ones, and MRA retains all consonants unchanged. This is evidently reflected in the $PC$ and $NC$ scores of these encoders, RI with $93.93$ ($PC$) and $79.72$ ($NC$), SX with $90.23$ ($PC$) and $88.79$ ($NC$), and MRA with $86.05$ ($PC$) and $95.36$ ($NC$). With the increasing degree of granularity from RI to MRA, the recall capability\footnote{It should be noted that the recall capability (intrinsic) measured by $PC$ refers to an encoder's ability to generate identical phonetic codes for corresponding ground-truth sounds under pairwise evaluation, while the recall capability (extrinsic) measured by the collision rate refers to an encoder’s tendency to assign the same representation to different inputs.} degrades as evidenced by decreasing $PC$ scores, while a corresponding improvement in precision is observed with increasing $NC$ scores. Notably, MRA also outperforms the English writing system (WS) in terms of $NC$, showing that simply removing vowels except those in word-initial position improves the overall precision of English words as phonetic representations. This result is consistent with findings from several previous studies indicating that vowels in fact account for the majority of letter-to-sound inconsistencies in English \cite{vowels1, vowels2}.

In the right subplot (b), the $AOC$ scores (y-axis) and the collision rates (x-axis) of each encoder are together summarized. The $AOC$ score reflects an encoder’s overall performance by capturing a reasonable trade-off between $PC$ and $NC$. The collision rate ($\mu$), on the other hand, corresponds to the average number of phonetic encoding matches and its optimal selection varies depending on the task. Thus, when collision rates are comparable, the encoder yielding a higher $AOC$ score is therefore preferable. As an illustration, PK performs optimally for tasks requiring a low collision rate, e.g., $1.5 < \mu < 2.5$, since it has the highest $AOC$ score in the given range. Similarly, MP, Phonix (PHIX) and AlphaSIS (ASIS) achieve optimal performance for $2.5 < \mu < 3.5$, $3.5 < \mu < 4.5$, and $4.5 < \mu < 5.5$, respectively. Among the remaining encoders with substantially high collision rates ($\mu >5.5$), SoundD (SDD) is preferable to RI due to its considerably higher $AOC$ score and comparable collision rate. Most notably, Phonex (PHEX) outperforms many other alternatives in terms of $AOC$ despite having the highest collision rate ($11.08$). A possible explanation for this behavior of PHEX may be attributed to its hybrid nature, combining features of MP followed by an encoding procedure similar to that of SX, which makes it a strong candidate for tasks that demand high collision rates. Within this analysis, PHEX stands out as the most intriguing case, achieving the highest extrinsic recall based on $\mu$ while maintaining a moderate level of intrinsic recall rate based on $PC$. Moreover, in the original paper, its recall performance was reported to be superior to those of MP and SX, but it showed slightly lower accuracy in a name matching task \cite{phonex}, which is in line with our observations.

The results for the encoders applied in German, French, and Swedish are given in Tables \ref{tab2}, \ref{tab3}, and \ref{tab4}, respectively. The provided values can be interpreted in the same way as their counterparts used for the English dataset. In German, Phonet yields the highest $AOC$ score of $75.83$, followed by Reth Schek $(70.39)$. Since the collision rates of these encoders are quite close to $1.0$, they can effectively be regarded as phonetic transcribers rather than codes. Although both outperform the $AOC$ score of the writing system $(69.89)$, Phonet exhibits superior $AOC$ compared to Reth Schek, thus making it a preferable choice in applications requiring a low collision rate such as transcription. On the other side, Haase $(44.87)$ is significantly outperformed by Kölner $(62.58)$ in terms of $AOC$, even though the latter has a collision rate $(2.49)$ that is approximately twice as high. Phonem offers a potentially preferable trade-off, combining improved $AOC$ $(68.18)$ compared to Kölner, with a reduced collision rate $(1.35)$. Taken together, Phonet, Phonem and Kölner can be considered the most suitable encoders for German, and the selection among them depends on the degree of collision rate required by the task. For French, Early Henry outperforms FONEM both in terms of collision rate and $AOC$, suggesting that the latter encoder can be preferable only in applications specifically demanding a low collision rate. Early Henry shows a similar peculiar behavior to PHEX (English) in that it avoids a loss in $AOC$ despite having a considerably high collision rate. In fact, its $AOC$ score $(73.01)$ surpasses that of the writing system $(59.08)$ by a substantial margin. Early Henry removes all vowels unlike FONEM and more effectively handles silent consonants, which could account for its improved $AOC$ score. As for Swedish, two encoders, i.e., SfinxBis and Wåhlin, were evaluated for performance comparison. Wåhlin yields a higher $AOC$ score $(81.35)$ than that of the writing system $(79.31)$ with a very low collision rate $(1.01)$. This result suggests that Wåhlin functions more as a phonetic transcriber, analogous to Phonet and Reth Schek. Thus, SfinxBis stands the only viable choice for tasks that require at least some degree of collision, although it comes at the cost of losing accuracy.

\begin{table}[t]
	\caption{Evaluation of the phonetic encoders in German}
	\label{tab2}
	\centering
	\setlength\tabcolsep{0.5em}
	\begin{tabular}{|c|c|c|c|c|c|c|}
	\hline
	\hline
	\textbf{Encoder} & $\boldsymbol{|\Sigma|}$ & $\boldsymbol{\mu}$ & $\boldsymbol{PC}$ & $\boldsymbol{NC}$ & $\boldsymbol{OC}$ & $\boldsymbol{AOC}$\\
	\hline
	\hline
	Writing System & 29 & 1.00 & 94.44 & 91.86 & 93.13 & 69.89\\
	\hline
	Haase & 9 & 1.65 & \textbf{98.97} & 78.69 & 87.67 & 44.87\\
	\hline
	Kölner & 9 & \textbf{2.49} & 95.38 & 88.17 & 91.63 & 62.58\\
	\hline
	Phonem & 15 & 1.35 & 94.84 & 90.86 & 92.81 & 68.18\\
	\hline
	Phonet & 25 & 1.04 & 95.00 & \textbf{93.99} & \textbf{94.49} & \textbf{75.83}\\
	\hline
	Reth Schek & 22 & 1.05 & 96.00 & 90.68 & 93.27 & 70.39\\
	\hline
	\end{tabular}
\end{table}

\begin{table}[t]
	\caption{Evaluation of the phonetic encoders in French}
	\label{tab3}
	\centering
	\setlength\tabcolsep{0.5em}
	\begin{tabular}{|c|c|c|c|c|c|c|}
	\hline
	\hline
	\textbf{Encoder} & $\boldsymbol{|\Sigma|}$ & $\boldsymbol{\mu}$ & $\boldsymbol{PC}$ & $\boldsymbol{NC}$ & $\boldsymbol{OC}$ & $\boldsymbol{AOC}$\\
	\hline
	\hline
	Writing System & 38 & 1.00 & 91.06 & 89.95 & 90.50 & 59.08\\
	\hline
	FONEM & 26 & 1.22 & \textbf{93.36} & 89.11 & 91.18 & 62.04\\
	\hline
	Early Henry & 22 & \textbf{8.42} & 89.84 & \textbf{98.05} & \textbf{93.76} & \textbf{73.01}\\
	\hline
	\end{tabular}
\end{table}

\begin{table}[t]
	\caption{Evaluation of the phonetic encoders in Swedish}
	\label{tab4}
	\centering
	\setlength\tabcolsep{0.5em}
	\begin{tabular}{|c|c|c|c|c|c|c|}
	\hline
	\hline
	\textbf{Encoder} & $\boldsymbol{|\Sigma|}$ & $\boldsymbol{\mu}$ & $\boldsymbol{PC}$ & $\boldsymbol{NC}$ & $\boldsymbol{OC}$ & $\boldsymbol{AOC}$\\
	\hline
	\hline
	Writing System & 29 & 1.00 & 94.30 & 95.96 & 95.12 & 79.31\\
	\hline
	SfinxBis & 18 & \textbf{1.74} & 92.45 & 91.51 & 91.98 & 65.83\\
	\hline
	Wåhlin & 27 & 1.01 & \textbf{95.14} & \textbf{96.07} & \textbf{95.60} & \textbf{81.35}\\
	\hline
	\end{tabular}
\end{table}

\begin{figure}[!t]
	\centering
	
	\includegraphics[width=0.75\textwidth]{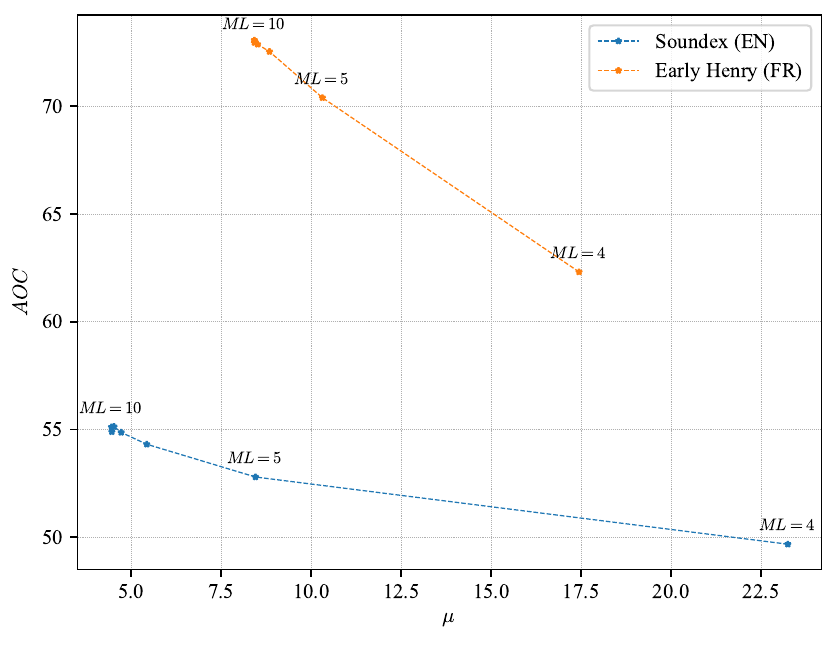}
	\caption{The $AOC$ scores (y-axis) and the collision rates (x-axis) of the encoders with varying maximum lengths}
	\label{fig2}
\end{figure}

Lastly, we conducted a separate experiment to investigate the impact of the maximum length (ML) parameter on $AOC$ scores and collision rates. When a phonetic code exceeds the specified ML parameter, the extra suffix part beyond that limit is removed. Modifying this parameter provides a simple yet effective means of regulating an encoder’s collision rate: lower ML limits yield higher collision rates, and vice versa. However, it is reasonable to expect that increasing it in this manner will come at the expense of losing the expressiveness of the phonetic code. This is clearly reflected in Figure \ref{fig2}, based on the results of the experiment carried out with two encoders, namely, SX and Early Henry, as the ML parameter was gradually increased. As seen from the figure, while the collision rate increases as the ML constraint becomes more strict, the $AOC$ score conversely decreases, especially more markedly for Early Henry than for SX. Accordingly, the observed decrease in the $AOC$ scores supports the expectation that truncated codes align less consistently with phonetic transcriptions.

\subsection{Orthographic Transparency}

\begin{figure}[!t]
	\centering
	\includegraphics[width=1.\textwidth]{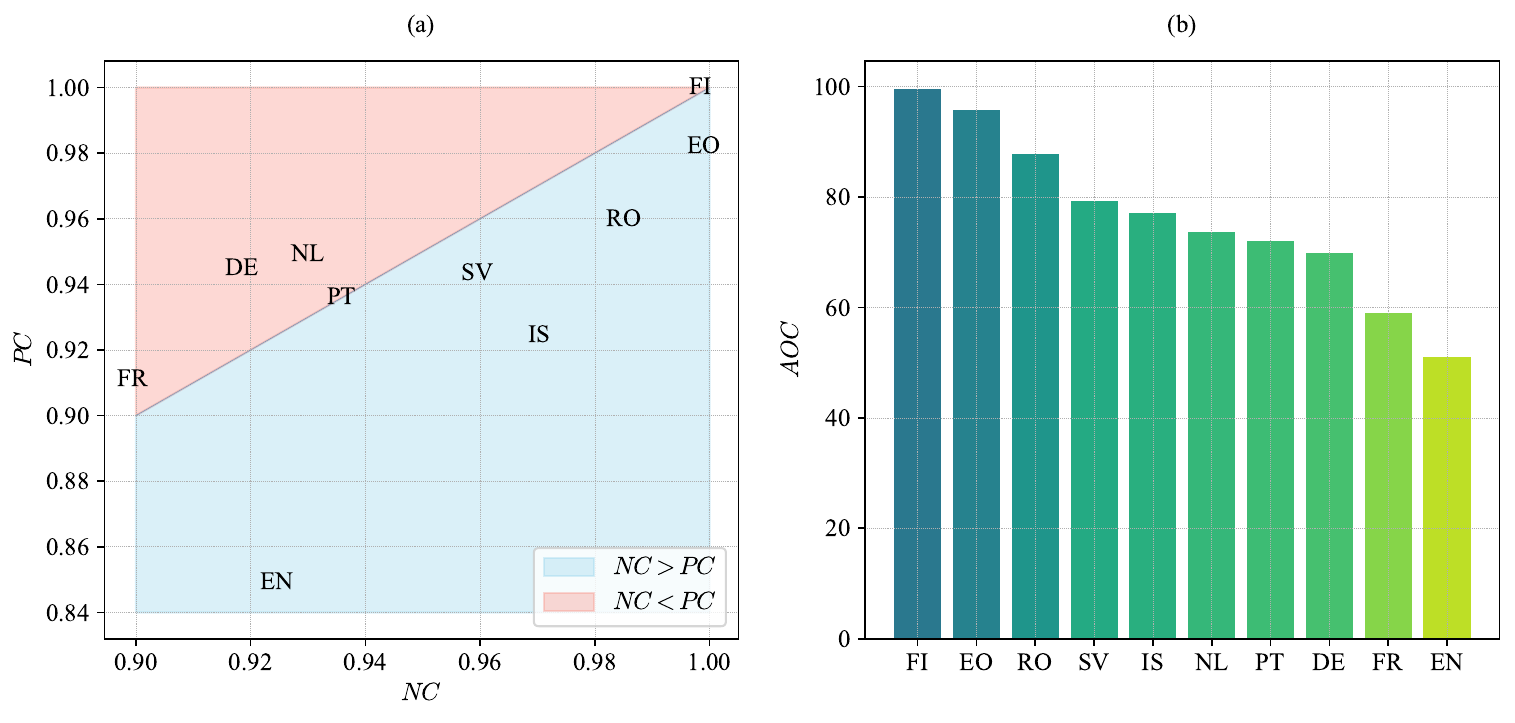}
	\caption{Evaluation of the writing systems of various languages indicating the degree of orthographic transparency. Provided are the $PC$ and $NC$ distribution (left) and the $AOC$ scores obtained for each language (right)}
	
	\label{fig3}
\end{figure}

In this section, we show that the $AOC$ score obtained by evaluating the written forms against the phonetic transcriptions can serve as a measure of orthographic transparency, which refers to the degree of how a language's orthography conforms to its spoken sounds \cite{ot0, ot1}. OTEANN is a seq2seq model developed to measure orthographic transparency through the model's decoding accuracy \cite{ot2}. It can operate in two directions, assessing reading transparency for grapheme-to-phoneme mapping and writing transparency for phoneme-to-grapheme mapping, whereas $AOC$ is symmetric in that it quantifies the level of consistency between written forms and transcriptions without taking directionality into account. In the left subplot (a) of Figure \ref{fig3}, the $PC$ and $NC$ scores are separately provided for 10 languages, i.e., Dutch (NL), English (EN), Esperanto (EO), Finnish (FI), French (FR), German (DE), Icelandic (IS), Portuguese (PT), Romanian (RO), and Swedish (SV). For languages that have higher $NC$ scores than $PC$, e.g., English and Icelandic (blue zone), false negatives are more predominant than false positives in pairwise concordance analysis, and vice versa (red zone). As shown by the AOC scores in the right subplot (b), Finnish has the most transparent orthography, while English has the most opaque. Finnish is recognized in the literature to be fully transparent \cite{ot3}, in contrast to notably opaque systems of English and French \cite{ot1}, aligning with our results. 

To conduct a comparative analysis, we also trained OTEANN to obtain both \textit{WRITE} and \textit{READ} accuracies using our dataset. For each language, $10,000$ samples were selected randomly and were then partitioned into 90\% for training and 10\% for testing. Table \ref{tab5} presents the results, with OTEANN's accuracy rankings determined with respect to the sorted $AOC$ scores, and accordingly, the coherence of the two approaches is quantified via the Kendall rank correlation coefficient \cite{tau} yielding the $\tau$ statistic of $0.46$ and $0.82$ for \textit{READ} and \textit{WRITE} accuracies, respectively. This implies that the AOC score aligns more strongly with OTEANN's performance in the writing task, yet it still exhibits a positive correlation with the reading task accuracy.

\begin{table}[b!]
	\begin{center}
		\caption{Orthographic transparency evaluation (AOC vs. OTEANN accuracy)}
		\centering
		\setlength\tabcolsep{0.5em}
		\begin{tabular}{|C{1.2cm}|C{0.75cm}|C{0.75cm}|C{0.75cm}|C{0.75cm}|C{0.75cm}|C{0.75cm}|C{0.75cm}|C{0.75cm}|C{0.75cm}|C{0.75cm}|}
			\hline
			\hline
			\textbf{Measure} & \textbf{FI} & \textbf{EO} & \textbf{RO} & \textbf{SV} & \textbf{IS} & \textbf{NL} & \textbf{PT} & \textbf{DE} & \textbf{FR} & \textbf{EN}\\
			\hline
			\hline
			AOC & 99.62 (1) & 95.77 (2) & 87.81 (3) & 79.31 (4) & 77.06 (5) & 73.67 (6) & 72.02 (7) & 69.89 (8) & 59.08 (9) & 51.08 (10)\\
			\hline
			OTEANN (\textbf{READ}) & 0.982 (2) & 0.988 (1) & 0.936 (3) & 0.714 (9) & 0.791 (6) & 0.777 (7) & 0.846 (5) & 0.771 (8) & 0.870 (4) & 0.428 (10)\\
			\hline
			OTEANN (\textbf{WRITE}) & 0.988 (2) & 0.989 (1) & 0.926 (3) & 0.754 (4) & 0.666 (6) & 0.639 (7) & 0.581 (8) & 0.697 (5) & 0.317 (9) & 0.264 (10)\\
			\hline
		\end{tabular}
		\label{tab5}
	\end{center}
\end{table}

\section{Conclusion and Future Work}

In this article, a novel evaluation scheme has been presented in order to measure the consistency of phonetic codes with phonetic transcriptions. A large number of phonetic encoders were evaluated across multiple languages based on the concordance measures, which were shown to serve as a fuzzy analogue to sensitivity, specificity and accuracy, together with their collision rates. The results were analyzed in detail to assess phonetic encoders with respect to their performance, highlighting their strengths and weaknesses that align with earlier observations in the literature. Furthermore, it was shown that the introduced measure may function as an indicator for orthographic transparency when applied to the written forms of a given language, yielding results consistent with those reported in previous studies.

The significance of the proposed approach arises from its ability to provide a task-independent framework where phonetic encoding algorithms can be evaluated based on their representational accuracy and recall capability without reliance on external evaluation. Several directions for future research can be addressed in order to extend the potential applicability of the framework. Considering that the final concordance score primarily depends on the distance metric used to measure pairwise similarities, incorporating additional distance metrics alongside the normalized edit distance as well as applying post-processing operations on similarity scores such as standardization could further improve generalizability across languages, particularly in the case of syllabic writing systems where one symbol may denote multiple sounds. In addition, employing various types of datasets such as those consisting exclusively of proper names may lead to different rankings among encoders in the evaluation, since a subset of these algorithms were originally designed for handling words of this nature.

\bibliographystyle{splncs03}

\bibliography{references}


%
%


\end{document}